%% file: acl2023.tex
\pdfoutput=1

\documentclass[11pt]{article}

\usepackage[]{ACL2023}

\usepackage{times}
\usepackage{latexsym}
\usepackage{gal_basic}

\usepackage[T1]{fontenc}

\usepackage[utf8]{inputenc}

\usepackage{inconsolata}

\title{Narrowing the Knowledge Evaluation Gap:\\Open-Domain Question Answering with Multi-Granularity Answers}

\author{Gal Yona \\
  Google Research \\
  \texttt{galyona@google.com} 
  \\\And
  Roee Aharoni \\
  Google Research \\
  \texttt{roeeaharoni@google.com}
  \\\And
  Mor Geva \\
 Tel Aviv University,\\Google Research \\
 \texttt{pipek@google.com}
  \\
  }

\usepackage{mdframed}
\mdfdefinestyle{style1}{
    backgroundcolor=yellow!05
}

\begin{document}

\maketitle
\begin{abstract}
\input{abstract}

\end{abstract}

\section{Introduction}
\label{section:intro}
Large language models (LLMs) often generate factual errors, especially when the task requires less widely-known knowledge \cite{mallen-etal-2023-trust, sciavolino2021simple}.
Such factual errors are commonly attributed to the LM lacking relevant knowledge \cite{zheng2023does} or over-committing to earlier mistakes \cite{zhang2023language}.

\begin{figure}[t]
\setlength{\belowcaptionskip}{-10pt}
    \centering
    \includegraphics[width=0.85\linewidth]{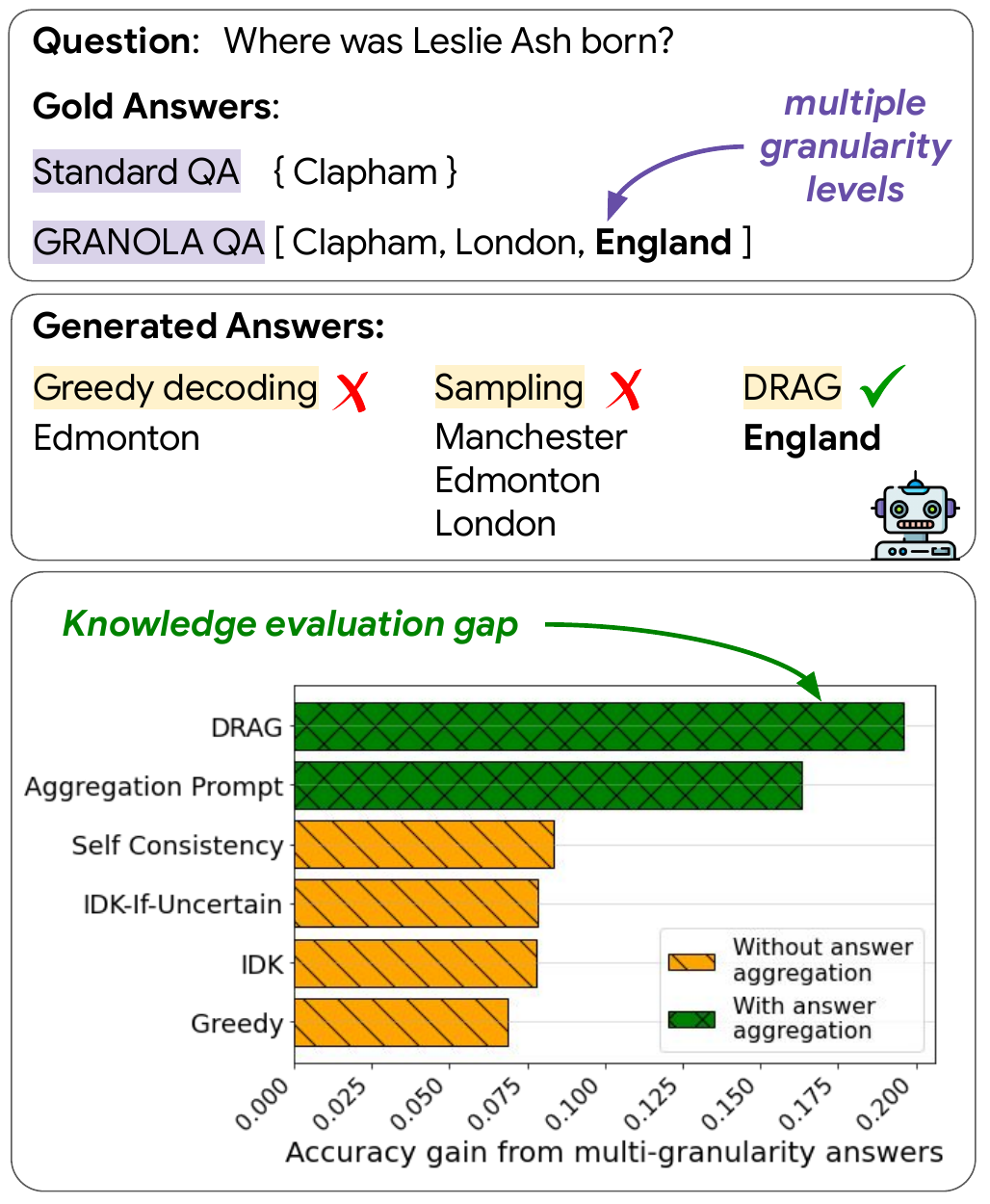}
    \caption{\textbf{Top}: \granola~QA evaluation with multi-granularity answers. \textbf{Middle}: Decoding with Response Aggregation (\drag) 
    outputs a (potentially coarser) response by
    aggregating several responses of the model. \textbf{Bottom}:
    Accuracy gain from evaluating 
    using multi-granularity answers for several decoding strategies.  
    \drag~reveals a significant knowledge evaluation gap. 
    }
    \label{fig:intro}
\end{figure}

We conjecture that factual mistakes can stem from a different failure source, when the model prioritizes different textual attributes (e.g., fluency or specific formats that appeared in the training corpora) over factuality. Such failures can result in generated text that mixes both correct and incorrect statements, even when the incorrect parts are not strictly required by the question.

Consider for example the question \nl{When was Mark Bils born?}. 
When prompting ChatGPT\footnote{Responses were obtained by querying ChatGPT 3.5 using the standard Web API in December 2023.}
for answering this question, sampled responses include \nl{March 22, 1958}, \nl{May 19, 1958} and \nl{August 15, 1958}. This may suggest that the model is confident that Bils was born in 1958 -- which is a correct answer in this case, albeit not the most informative one -- yet it displays a preference for outputting a more detailed but incorrect response in a specific full-date format.

This example also highlights how factual questions can be answered correctly \textit{at different levels of granularity}. Namely, while the answers \nl{December 1, 1958}, \nl{December 1958}, and \nl{1958} vary in terms of informativeness, they are all factually correct. However, answer granularity levels are not considered in standard question answering (QA) settings, which typically evaluate a predicted answer based on its similarity to a set of reference answers of the same (usually the most-specific) granularity. Even when different levels of granularity are present, there is no notion in which matching to a more specific answer is ``better''. As a result, standard QA evaluation may significantly \emph{underestimate} the knowledge encapsulated in LMs, a phenomenon which we refer to as the \emph{knowledge evaluation gap}. Indeed, recent human evaluation suggests that such granularity disparities account for approximately 10-15\% of the disagreements between lexical matching and human evaluation \cite{kamalloo2023evaluating, zheng2023does}.

In this work, we tackle this problem by proposing a novel multi-granularity QA evaluation setting, called \granola~QA (short for \textbf{GRAN}ularity \textbf{O}f \textbf{LA}bels). Unlike existing evaluation, in \granola~QA questions are labeled with ground-truth answers with multiple levels of granularity and predicted answers are evaluated in terms of both their \textit{accuracy} and \textit{informativeness} (\S\ref{section:granola_qa}). The evaluation is done using two new metrics: \granola~Accuracy, which checks if there was a match against \emph{any} of the answers, and \granola~informativeness, which is a weighted score prioritizing fine-grained correct answers over their coarse-grained counterparts.

Next, we present a simple and general methodology for augmenting an existing single-granularity QA dataset to the setting of \granola~QA, which does not involve any human labor (\S\ref{section:enriching}). This process is based on obtaining additional information about entities present in the original questions and answer(s) from an external knowledge graph (KG), and then using an LLM to form multi-granularity answers conditioned on this information. We apply our methodology on the \abbreviatedeq{} dataset~\cite{sciavolino2021simple}, using WikiData \cite{vrandevcic2014wikidata} as the KG. The resulting dataset, \graneq, consists of 12K QA examples with an average of 2.9 multi-granularity answers per question. A manual analysis of a random subset of the data shows that our automatic procedure yields highly-accurate answers.

We evaluate various baselines on \graneq{}, including greedy decoding and  
methods that abstain from answering in cases of uncertainty
\cite{yoshikawa-okazaki-2023-selective, yang2023uncertainty, yang2023alignment, ren2023self}. In addition, we introduce a novel decoding strategy, called Decoding with Response Aggregation (\drag), that is geared towards aligning the granularity level of a model's response with its uncertainty level (\S\ref{section:drag}). \drag~uses temperature sampling to obtain a set of candidate responses, and then answers the original question based on \emph{an aggregation of these responses}, which we implement using few-shot prompting. Figure~\ref{fig:intro} depicts an example of \drag{}'s aggregation of several incorrect responses into a correct coarser answer that matches against the multi-granularity labels.

Our experiments (\S\ref{section:experiments}) show that: (1) with standard decoding the gap between \granola~accuracy and standard accuracy is small, which corroborates that LMs tend to output detailed responses, even when these are incorrect, (2) with \drag{} this gap is high, showing that unlike standard decoding, \drag{} outputs coarse answers, (3) \granola{}~accuracy remains high with \drag{} even for rare entities, 
suggesting that LLMs know less \textit{detailed} information about them rather than lacking any knowledge \cite{mallen-etal-2023-trust},
(4) compared to standard decoding and methods that allow the model to abstain from answering (``IDK''), \drag~yields a better trade-off between factuality and response informativeness, and (5) this evaluation gap is not observed when using semantic similarity scores against single-granularity reference answers.

To summarize, 
we introduce \granola{}, a new QA evaluation setting that considers both the accuracy and informativeness of predicted answers. We propose a simple automatic procedure for generating accurate multi-granular answers for given QA pairs, and apply it to the \eq{} dataset to create \graneq{}. We introduce a new decoding scheme, called \drag{}, tailored to modify the response to a level of granularity that fits the model's uncertainty levels. We show that \drag{} improves both informativeness and accuracy (relative to standard decoding), and that standard evaluation may significantly under-estimate the knowledge of LMs, especially about rare entities.

\section{\granola{} Question Answering}
\label{section:granola_qa}

We formalize the setting of \granola~QA and define new metrics for quantifying accuracy and informativeness of QA predictions.

\subsection{Problem Setting}

In a typical open-domain QA setting \cite{yang2015wikiqa, voorhees1999trec, kwiatkowski2019natural, joshi2017triviaqa, sciavolino2021simple},
a model predicts an answer $p$ to a given question $q$, which is evaluated against an unordered set of gold answers $\mathcal{A} = \set{a_1, \dots, a_k}$. 
The evaluation usually relies on lexical matching with standard metrics like exact-match or token-F1 between the predicted answer and each of the gold answers.\footnote{The answers are typically being normalized (i.e. case-folding and removing punctuation and articles).} For example, a possible set of answers to the question \nl{Where is the headquarter of Guildhall School of Music and Drama?} would be $\set{\text{Barbican Centre}, \text{The Barbican}}$. Importantly, the gold answers in $\mathcal{A}$ are interchangeable, where matching against either of $a_1$ or $a_2$ is equally good.

However, 
we observe that a question may be answered correctly at different levels of granularity. Namely, ``London'' is also a correct answer to the question, since the Barbican Centre is located there. If ``London'' does not appear in  $\mathcal{A}$, standard evaluation will render this answer as incorrect, resulting in under-estimating the LM's knowledge. Moreover, if London is included in  $\mathcal{A}$, then answering either ``London'' or ``The Barbican'' is considered equally correct, despite the fact that the second answer is more specific and arguably more valuable. 

Here we propose that QA predictions should be evaluated while considering different granularity levels, a setting which we name \granola{} QA. Formally, the answer $p$ should be evaluated against an \emph{ordered set of multi-granular} gold answers $\hat{\mathcal{A}} = \set{ \mathcal{A}_1, \dots, \mathcal{A}_\ell}$. Here, $\mathcal{A}_1$ is the set of the most informative correct answers (e.g. $\set{\text{Barbican Centre}, \text{The Barbican}}$) and $\mathcal{A}_\ell$ is the set of least-informative correct answers (e.g. ``London'' could be in $\mathcal{A}_2$ and ``UK'' in $\mathcal{A}_3$).

\subsection{Evaluation}

At a high-level, we will evaluate \granola~QA performance across two 
axes: \emph{accuracy} and \emph{informativeness}. Accuracy is determined based on whether the candidate answer matches against \emph{any} of the \granola~answers; informativeness will reward matching against fine-grained answers by using an appropriate weighting scheme:

\begin{definition}[\granola~Evaluation]
\label{def:granola_metrics}
Given a question $q$, an answer $p$ and \granola~labels $\hat{\mathcal{A}}$, accuracy and informativeness are evaluated based on a simple two-step procedure:

\textbf{Step 1: Find a match.} Let $i^\star \equiv i^\star(p; q, \hat{\mathcal{A}})$ denote the smallest index $i \in [k]$ for which there is a match between $p$ and $\mathcal{A}_i \in  \hat{\mathcal{A}}$ (meaning the F1 score between $p$ and an answer in $\mathcal{A}_i$ exceeds some threshold $\tau$), or $\perp$ if no match is found.

\textbf{Step 2: Evaluate.} \granola~accuracy is defined as $\mathbf{1}[i^\star \neq \perp]$. Informativeness is defined as $\exp(-\lambda \cdot (i^\star-1))$, or $0$ if no match was found.

\end{definition}

The notion of informativeness relies on a weighting scheme 
that assigns a weight of $1.0$ to the fine-grained answers $\mathcal{A}_1$, and exponentially decreasing weight for answers $\mathcal{A}_{i>1}$. This represents the diminished utility of coarser answers. The parameter $\lambda$ can be used to control the rate of decrease: as $\lambda \to 0$ coarser answers receive higher weights; see Appendix~\ref{appendix:figures} for a visualization of how the weights behave as a function of $\lambda$.

\section{Enriching QA Samples with Multi-Granularity Answers}
\label{section:enriching}

We turn to the question of constructing \granola~QA benchmarks.
We observe that multi-granularity answers are in principle \textit{abstractions} of the most-detailed answer. For example (see Figure 3), the answer ``Michael Madhusudan Dutta'' to the question \nl{Who translated the play Neel Darpan into English?} can be abstracted into a higher-level description such as ``An Indian Poet''. Therefore, one way to generate multi-granularity answers is to start from an existing QA pair
and enriching it with multi-granularity answers through abstraction.

Following this approach, we describe a simple and automatic procedure for adjusting factual QA datasets to \granola~QA (\S\ref{section:automatic}).
Then, we apply this procedure to the \eq~dataset (\S\ref{section:granola_eq}), a widely used entity-centric QA dataset \cite{sciavolino2021simple}, to create a multi-granularity QA benchmark. Last, we manually analyze the quality of the generated data (\S\ref{section:data_quality}).

\begin{figure}[t]
\setlength{\belowcaptionskip}{-10pt}
    \centering
    \includegraphics[scale=0.32]{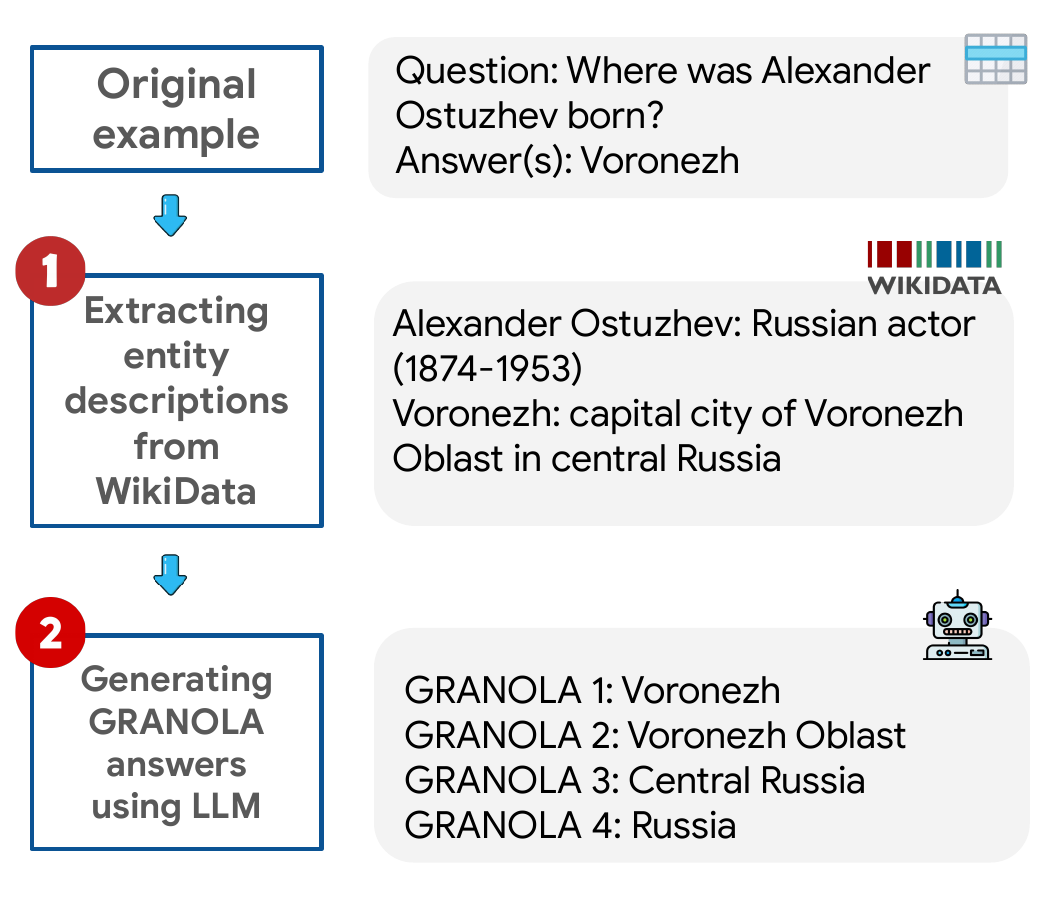}
    \caption{Our procedure for adding multi-granularity answers to given QA pairs.
    }
    \label{fig:granola-eq-overview}
\end{figure}

\subsection{Automatic Answer Generation}
\label{section:automatic}

\begin{figure}[t]
\setlength{\belowcaptionskip}{-8pt}
    \centering
    \includegraphics[width=0.99\linewidth]{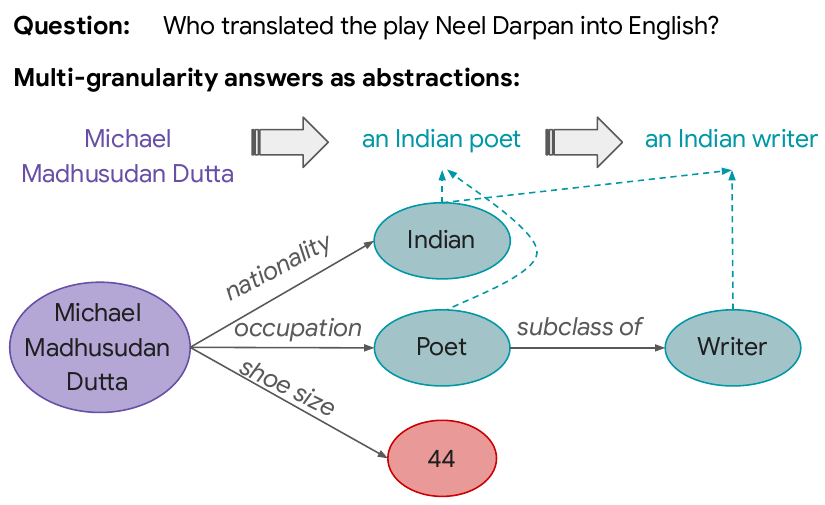} 
    \caption{ 
    An illustration of multi-granularity answers as entity abstractions. Given an answer entity, we use an external KG to generate coarser answers from its properties (\textcolor{teal}{turquoise}) in addition to the original answer  (\textcolor{violet}{purple}). Notably, not all KG properties are equally good candidates for multi-granular answers  (\textcolor{red}{red}).
    }
    \label{fig:abstractions}
\end{figure}

We focus on evaluating factual knowledge in LLMs, where the answer to a given question is an entity (e.g., a person or a place). Given an answer, we propose to generate coarser versions of it by utilizing an external knowledge graph (KG). Specifically, given a KG with facts encoded as subject-relation-object triplets (e.g., the triplet $(\texttt{Paris}, \texttt{capital of}, \texttt{France})$ would encode the fact that Paris is the capital of France) and an answer entity $e$, coarser versions of $e$ can be obtained by replacing it with higher-level properties of it in the KG.
For example (Figure~\ref{fig:abstractions}), replacing the answer ``Michael Madhusudan Dutta'' with its properties of \texttt{Nationality} and \texttt{Occupation} would create a new coarser answer ``Indian Poet''.

In principle, however, there are many possible answer properties that can be used -- and intuitively, not all of them are key properties of the entity that are useful for evaluating general factual knowledge. For example, answering the original question with Michael Madhusudan Dutta's shoe size is not what we want to capture by coarse answers. Thus, to create a generic methodology for enriching an existing QA dataset with answers, we must be able to automatically determine the relevant properties.

To overcome this challenge, instead of relying on KG triplets directly, we use short textual descriptions that capture the key properties of the entity in the KG. Such descriptions are often offered by knowledge sources such as WikiData. 
For example, the entity Michael Madhusudan Dutta has the following description:  \nl{Bengali poet and dramatist}.

Overall, our answer generation process has two steps, depicted in Figure~\ref{fig:granola-eq-overview}. Given a QA pair, we first obtain a description of the answer entity and any entities appearing in the question from an external KG. Then,   
we zero-shot prompt an LLM to generate an ordered list of answers at varying levels of granularity, conditioned on the given QA pair and the entity descriptions. See Table~\ref{tab:proc_prompts} for the exact instruction prompt.

\subsection{\granola~\eq}
\label{section:granola_eq}

We apply the procedure described in \S\ref{section:automatic} to enrich the test split of  
\abbreviatedeq~\cite{sciavolino2021simple} with \granola~answers. \eq~is an entity-rich QA dataset, created by 
converting factual subject-relation-object triples into natural language questions using manually-defined templates. We use PaLM 2-L as the LLM \cite{anil2023palm}.

The resulting dataset, which we refer to as \graneq, spans 16 relations and has a total of 12,452 examples. Overall, our procedure yielded 2-3 coarser answers per questions 
($\sim$20\% have 2 answers overall, $\sim$60\% have 3, and $\sim$15\%  have 4 or more; this is distributed relatively uniformly over relations). 
Examples from \graneq{} are shown in Table~\ref{table:granola_samples}. More details are in Appendix \ref{appendix:granola-eq}.

\subsection{Data Quality}
\label{section:data_quality}

We manually evaluate the quality of a generated answer $a$ with respect to a question $q$ from \graneq{} across the following axes:

\begin{itemize}
[itemsep=1pt, topsep=2pt,leftmargin=*]
    \item \textbf{Correctness}: We use WikiData to verify whether  $a$ is a factually correct answer to $q$. 
    Notably, while $a$ was generated conditioned on the description, the LLM might produced it while relying on its parametric knowledge rather the information in the description.
    For example, for the question \nl{Where did Marcel Gaumont die?}, the model generated the answers ``Paris'', ``Île-de-France'', and ``France'' while the WikiData description of Paris is \nl{Capital of France}. Therefore, in this case the LLM used its parametric knowledge to add a new granularity level (Île-de-France).
    \item \textbf{Informativeness}: We verify that $a$ is a non-trivial answer to $q$. We consider an answer as trivial if it could be generated based on the question template alone (i.e., a version of $q$ in which the entity is redacted). For example, \nl{Earth} is a trivial answer to the question \nl{Where was Fiona Lewis born?} because it could be obtained based on the template \texttt{Where was [X] born?}.
    \item \textbf{Granularity}: We assess whether $a$ is coarser than the answers preceding it. 
    For the first \granola~answer, we define this as whether the answer is identical to the original answer.
\end{itemize}

\begin{table}[t]
\setlength{\belowcaptionskip}{-10pt}
\centering
\footnotesize
\begin{tabular}{p{0.42\linewidth}p{0.45\linewidth}}
\textbf{Question} & \textbf{\granola~Answers} \\
\toprule
\nl{Where was Fiona Lewis born?} & Westcliff-on-Sea; Essex; England \\
\midrule
\nl{What music label is Courage represented by?} & Rock Records; a Taiwanese record label \\
\midrule
\nl{Who is August von Hayek's child?} & Friedrich Hayek; an economist \\
\midrule
\nl{Who is the author of The Adding Machine?} & Elmer Rice; an American playwright; a playwright \\
\midrule
\nl{Where was Toby Shapshak educated?} & Rhodes University; Makhanda, South Africa; South Africa \\
\bottomrule
\end{tabular}
\caption{Examples from \graneq. Answers are separated by a semicolon and listed fine-to-coarse. The first answer is the original answer in \eq; subsequent answers were generated (see \S\ref{section:automatic}).}
\label{table:granola_samples}
\end{table}

We treat these metrics as binary and manually evaluate a sample of 1\% of the data (124 questions and their corresponding 358 answers). Table~\ref{table:granola_human_eval} reports the fraction of examples in each error category with a representative example. Our evaluation reveals that the enriched answers are of high-quality, with 
all of the generated answers being factually correct 
Nonetheless, there is headroom for improving our answer generation procedure. E.g., there are examples with useful information in the description that is not utilized by the model, (suggesting the knowledge evaluation gap may be even larger than observed in our results in \S\ref{section:experiments}.)

\begin{table}[t]
\setlength{\belowcaptionskip}{-10pt}
\footnotesize
\centering
\begin{tabular}{p{0.22\columnwidth}p{0.67\columnwidth}}
 \textbf{Error type (\%)} & \textbf{Example} \\ \toprule
Informativeness (6\%) & \underline{Question:} What music label is Sarah Buxton represented by? \underline{Answers:} Lyric Street Records; a music label \\ \midrule
Granularity (9\%) & \underline{Question:} Who owns Eccles Coliseum? \underline{Answers:} Southern Utah University; a public university; a public university in Utah \\ \bottomrule
\end{tabular}
\caption{Human evaluation results of \graneq, showing for each error type the fraction of erroneous cases and an example.  
}
\label{table:granola_human_eval}
\end{table}

\section{Decoding with Response Aggregation}
\label{section:drag}

Humans naturally tailor the granularity level of their responses to their uncertainty levels.
Consider asking a person A, when another person B was born. The format of the response will depend on the relationship between A and B, and specifically on how much A knows about B. For example, if A is extremely familiar with B (e.g., B is A's son), then we expect the answer to include the full date of birth. If A is only partially familiar with B (e.g., B is a celebrity that A knows), then we expect the answer to be more generic (e.g. only the year or decade). If A is not familiar with B, then we expect A to say that they do not know the answer.

In this section, we propose a novel decoding strategy, called Decoding with Response Aggregation (\drag), that is intended to encourage LMs to do the same. We focus on a fixed (i.e., frozen) LM, and our objective is to improve factuality at inference time by attempting to provide a coarser answer in the place of a fine-grained but incorrect answer. In \S\ref{section:experiments}, we will evaluate our proposed decoding strategy against various existing baselines on the \granola~QA dataset we constructed.

\drag{} consists of two stages: 
\begin{itemize}
[itemsep=1pt, topsep=2pt,leftmargin=*]
    \item \textbf{Sampling}: We sample $N$ responses from the model with temperature $T>0$.
    \item \textbf{Aggregation}: The final output is 
    the most informative response that is consistent with the set of sampled responses. This can be implemented in different ways, e.g. via prompting an LLM.
\end{itemize}

Revisiting the example question \nl{When was Mark Bils born?} (\S\ref{section:intro}), aggregating the sampled responses \nl{March 22, 1958}, \nl{May 19, 1958} and \nl{August 15, 1958}, 
should yield \nl{1958}. Pseudo-code for \drag{} is provided in Figure~\ref{algo:drag}.

\paragraph{Choice of hyperparameters} The sampling temperature $T$ and number of responses $N$ control the trade-off between factuality and informativeness. Intuitively, larger values of $T$ and $N$ encourage more diverse outputs and hence more aggregation, favouring factuality over informativeness. 

\paragraph{\drag~vs existing decoding strategies} When $N=1$, the aggregation is trivial and \drag~recovers standard decoding strategies (e.g. greedy decoding or temperature sampling, based on the value of $T$). Conceptually, \drag~is also a generalization of other popular decoding strategies that are based on sampling a set of candidate responses. For example, replacing our proposed aggregator with a naive aggregation that outputs the majority response
recovers \emph{self-consistency} \cite{wang-etal-2022-self}.

\begin{figure}[t]
\setlength{\belowcaptionskip}{-8pt}
    \centering
    \small
    \begin{mdframed}[style=style1]
      \begin{algorithmic}
      \STATE {\bfseries Hyperparameters:} Temperature $T > 0$; number of samples $N$
        \STATE {\bfseries Input:}  Input $x$; Model $M$
        \vspace{1mm}
        \STATE Generate $\set{r_1,\dots, r_N}$ continuations for $M(x)$ at temperature $T$;
        \STATE Let $\hat{r} = \ra\br{\set{r_1,\dots, r_N}}$;
        \RETURN The aggregated response $\hat{r}$
        \vspace{1mm}
      \end{algorithmic}
\end{mdframed}
    \caption{Decoding with Response Aggregation (\drag). We implement \ra{} by instructing an LLM to output what $r_1, \dots, r_N$ have in common, or IDK if they do not share meaningful properties.
    }
    \label{algo:drag}
\end{figure}

\section{Experiments}
\label{section:experiments}

We assess how accounting for answer granularity, both in evaluation 
and during decoding, 
influences the evaluation of LLM performance on factual questions. After describing our experimental setting (\S\ref{section:experimental_setting}), we compare between evaluation with standard accuracy and \granola~accuracy (\S\ref{section:experiments:knowledge_gap}), which reveals that current QA settings underestimate LLMs' knowledge. Then, we show that the gains in accuracy from using \granola{} cannot be matched by existing semantic similarity scores (\S\ref{section:drag_evaluation}), which highlights the utility of this setting in capturing differences between multi-granularity answers.
Last, we use the \granola~metrics to evaluate \drag{} with respect to baselines in terms of accuracy and informativeness (\S\ref{section:drag_evaluation}), showing its superiority in decoding answers that are tuned towards the LLM's knowledge.

\subsection{Experimental Setting}
\label{section:experimental_setting}

We evaluate \drag{} and multiple baselines on \graneq{} in a closed-book setting, where factual questions must be answered without access to an external knowledge source \cite{petroni2019language}.

For the aggregation stage of \drag{}, we instruct an \emph{aggregator} LLM to output what the sampled responses have in common or IDK if the responses have nothing meaningful in common (see Table~\ref{tab:proc_prompts} in Appendix~\ref{appendix:prompts} for the exact prompt). 

\paragraph{Baselines} We consider the following methods:

\begin{itemize}
[itemsep=1pt, topsep=2pt,leftmargin=*]
    \item \textbf{Standard Decoding}: We evaluated both greedy decoding (\greedy) and temperature sampling (\ts), but since \ts{} consistently under-performed \greedy{} we report results only for \greedy{}. 

    \item \textbf{I don't know (IDK)}: Given the established success of steering model behaviour via prompting \cite{mishra2021reframing, si2022prompting, ganguli2023capacity}, we consider two prompt-based IDK variants. In \IDK, the model is instructed to either answer the question or output IDK. In \IDKifuncertain, the model is specifically instructed to output IDK if its uncertainty is high.
    
    \item  \textbf{Aggregation-based baselines}: We evaluate \drag{} and 
    \IDKwithagg, in which we instruct the model to answer at a level of granularity that matches its uncertainty. As an ablation for the importance of the aggregation step in \drag{} we also evaluate 
    \selfconsistency~\cite{wang-etal-2022-self}, where we sample $N$ responses at temperature $T$ and output the majority response.\footnote{After 
    case-folding and removing punctuation and articles.}
    As noted in \S\ref{section:drag}, \selfconsistency~can be cast as an instance of \drag~with a simple aggregator (majority rule). 
\end{itemize}
See Table \ref{tab:algo_prompts} for the prompts used for the baselines.

\paragraph{Evaluation} We use \emph{\granola{} accuracy} and \emph{informativeness} as described in Definition \ref{def:granola_metrics}. 
To account for cases of IDK predictions, 
we adopt the perspective of \emph{selective prediction} \cite{el2010foundations, geifman2017selective} with recent applications in QA \cite{kamath2020selective} and text generation \cite{yoshikawa-okazaki-2023-selective}. Informativeness is left as is, except that IDK predictions are defined to contribute a score of $0.0$, since they are not informative at all. \granola{} Accuracy is replaced with  \emph{selective \granola{} accuracy}, which is the mean \granola~accuracy on the subset of predictions which are not IDK.

\paragraph{Models} We use instruction-tuned versions of PaLM 2-M and PaLM 2-L, the medium and large variants of the PaLM 2 LLM  \cite{anil2023palm}.

\begin{figure}[t]
    \centering
    \includegraphics[width=0.99\linewidth]{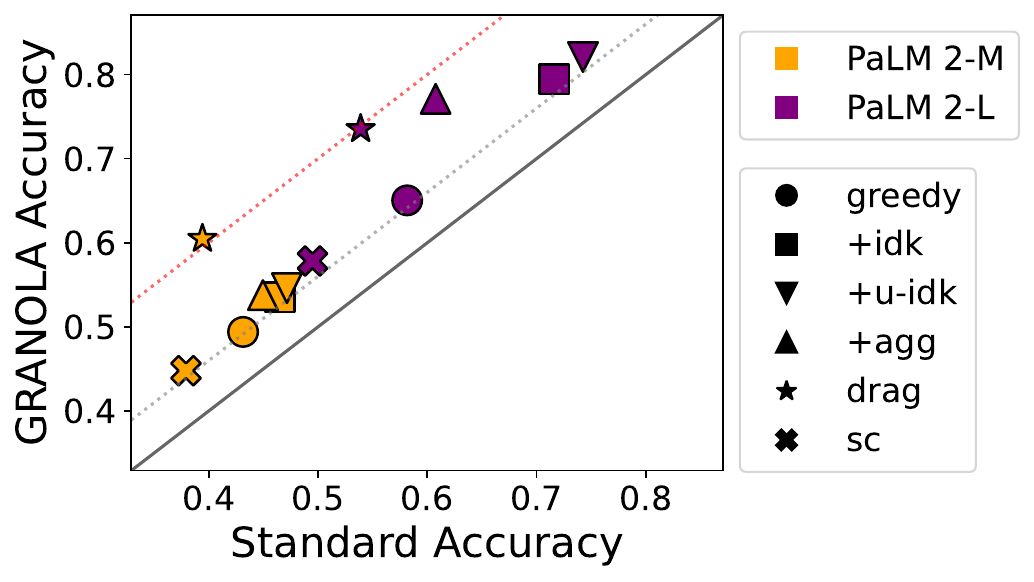}
    \caption{Standard accuracy vs. GRANOLA accuracy for the different models we evaluate.
    }
    \label{fig:acc_vs_granola_acc}
\end{figure}

\begin{figure}[t]
\setlength{\belowcaptionskip}{-10pt}
    \centering
    \includegraphics[width=0.4\textwidth]{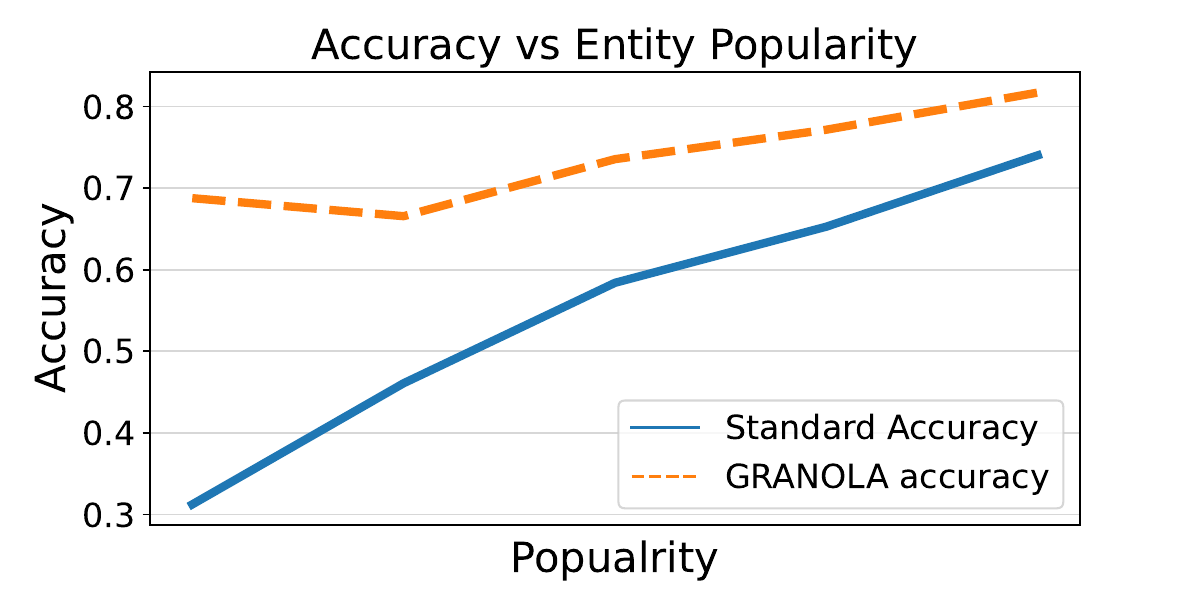}
    \caption{Accuracy vs. entity popularity for PaLM 2-L using \drag{}. Unlike standard accuracy, which declines steeply in popularity, \granola~accuracy plateaus.
    }
    \label{fig:acc_vs_pop_24b}
\end{figure}

\subsection{Knowledge Evaluation Gap}
\label{section:experiments:knowledge_gap}

Figure~\ref{fig:acc_vs_granola_acc} shows \granola~accuracy as a function of standard accuracy, for the different models and methods.
Note that the vertical distance from the $x=y$ line (black) represents the gain in accuracy from evaluating using multi-granularity answers. 
We observe that this gap is similar and relatively small of $\sim$5 points (grey dotted line) for methods that do not explicitly incorporate aggregation. This confirms our initial conjecture that standard decoding tends to generate detailed but incorrect responses. In addition, for the aggregation methods, this gap is substantially larger, nearing a $\sim$20 point increase (red dotted line). This demonstrates that both explicit aggregation (\drag) and implicit aggregation obtained via prompting can successfully steer the model towards tailoring its response granularity. It also reveals that the knowledge evaluation gap is both a function of existing evaluation practices \emph{and} standard decoding strategies. 
In Figure \ref{fig:acc_vs_granola_acc_by_relation} in Appendix \ref{appendix:results} we show a breakdown of these results to the different relations in \graneq, revealing that certain relations 
especially gain from multi-granularity answers.

Next, we consider how this gap behaves as a function of \emph{popularity}.\footnote{We quantify popularity using Wikipedia page-views for the question entity's page.} In Figure~\ref{fig:acc_vs_pop_24b} we stratify \graneq~into equally sized bins by popularity (x-axis) and compare standard accuracy (blue) with \granola~accuracy (orange, dashed). While standard accuracy steeply declines with popularity, \granola~accuracy plateaus. This reveals that models do capture knowledge about even very rare entities (but this knowledge is coarser). In Figure~\ref{fig:acc_vs_pop_340b} (\S\ref{appendix:figures}) we show that this behaviour is not demonstrated by standard decoding.

\subsection{Evaluation of \drag{}}
\label{section:drag_evaluation}

\begin{figure}[t]
\setlength{\belowcaptionskip}{-8pt}
    \centering
    \includegraphics[width=0.99\linewidth]{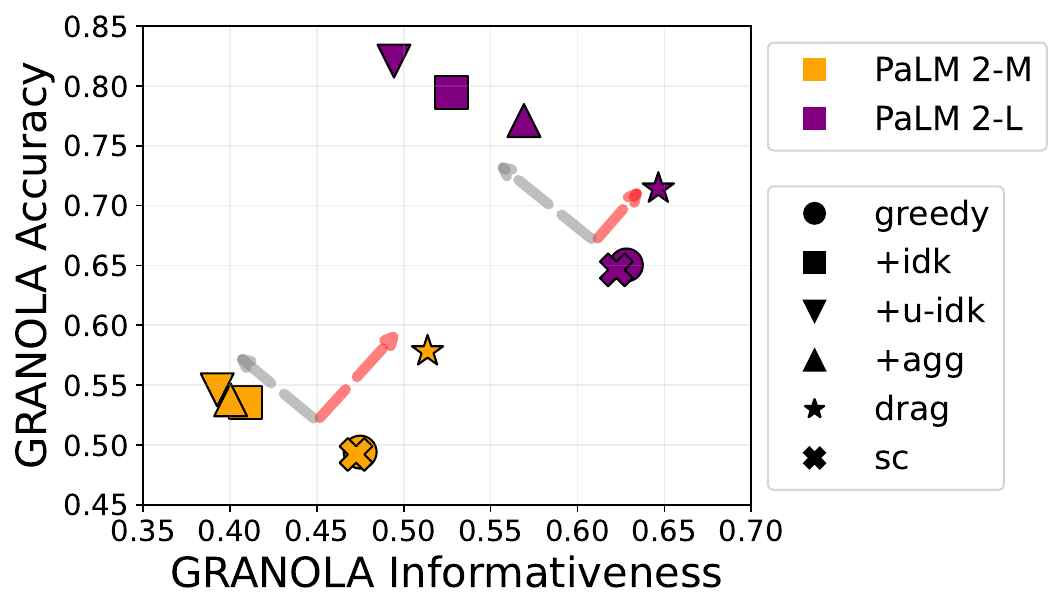}
    \caption{Answer accuracy vs. informativeness when using \drag{} compared to the baselines. Behaviour is consistent across model sizes (purple/orange): IDK baselines improve accuracy at the cost of making less informative predictions (grey arrow); \drag~improves both accuracy and informativeness (red arrow).}
    \label{fig:drag_eval}
\end{figure}

\begin{figure}[t]
\setlength{\belowcaptionskip}{-8pt}
    \centering
    \includegraphics[width=0.8\linewidth]{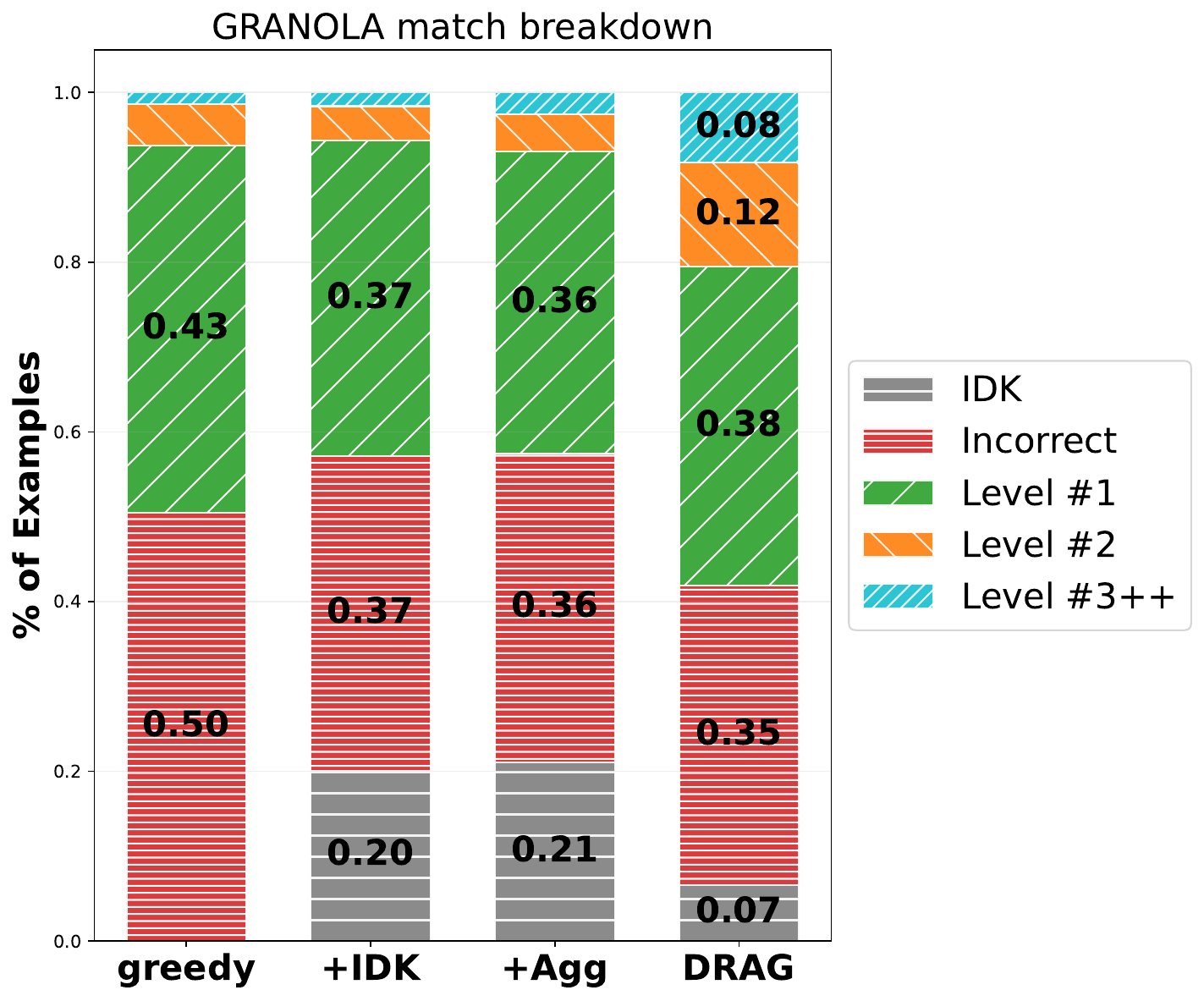}
    \caption{The granularity of answers predicted by PaLM 2-M. Level numbers correspond to the answer index in the ordered set of \granola{} answers, with 1 being the most fine-grained. While all methods decrease the fraction of errors compared to greedy (from 50\% to 35-37\%; \textcolor{red}{\textbf{red}}), \drag{} does this with significantly fewer IDK predictions (e.g., 7\% vs 20\%); \textcolor{gray}{\textbf{gray}}) and more coarse correct answers (e.g. 20\% vs 5\%). 
    }
    \label{fig:granola_match_24b}
\end{figure}

Figure~\ref{fig:drag_eval} shows the \granola~accuracy and informativeness of \drag{} compared to the baselines. The results are consistent across model sizes (purple vs orange). Figure~\ref{fig:granola_match_24b} provides a more detailed picture of the distribution of which \granola{} answer matched against the predicted answers (see Definition \ref{def:granola_metrics}). 
We distill several key takeaways:

\textbf{(1)} IDK baselines improve accuracy at the cost of less informative predictions (grey arrows in Figure~\ref{fig:drag_eval}): As expected, abstention (IDK) improves the selective accuracy. However, as evident in Figure~\ref{fig:drag_eval}, this comes at the cost of predictions that are overall less informative. For example, the fraction of errors made by \IDK{} drops from 42\% to 31\% -- but 17\% of the predictions are IDK. The number of coarse correct answers is unchanged at $\sim$5\%.

\textbf{(2)}  \drag{} improves both accuracy and informativeness (red arrows in Figure~\ref{fig:drag_eval}): Compared to standard decoding, \drag{} improves both accuracy and informativeness. As evident from Figure~\ref{fig:drag_eval}, this is obtained by a smaller fraction of abstentions (6\%) and a significantly larger fraction of coarse correct answers (16\%). This result confirms our original conjecture that the dichotomy (know/don't know) underlying IDK methods is too coarse.

\subsection{Meta-evaluation}
\label{section:results:meta_evaluation}
In the previous sections, we showed that multi-granularity answers facilitate a more faithful evaluation of LLM performance on factual questions. 
Here, we check whether a similar effect could be obtained by evaluating with semantic similarity against single-granularity reference answers.

To this end, we test if semantic similarity against single-granularity answers can distinguish between answers that \granola{} accuracy deems correct and incorrect. Concretely, we stratify \graneq{} according to whether both the standard and \granola~F1 scores exceed a threshold $\tau$, and report the mean semantic similarity score for each of the four resulting subsets. Note that, by definition, the standard F1 is a lower bound to \granola{} F1, so one of the subsets is empty.

Table~\ref{table:bleurt_scores} shows the results when using BLEURT \cite{sellam2020bleurt} as the semantic similarity metric. The mean BLEURT score is similar for examples that are incorrect according to both metrics and for examples that are correct only according to \granola~accuracy (gray rows). This highlights that BLEURT is not a good proxy for matching against multi-granularity answers.
Examples from \graneq~where \granola~accuracy disagrees with both standard accuracy and BLEURT score are provided in Table~\ref{tab:examples_mismatch} (Appendix~\ref{appendix:results}).

\section{Related work}
\label{section:related}



\newcolumntype{C}[1]{>{\centering\arraybackslash}p{#1}}
\begin{table}[t]
\setlength{\belowcaptionskip}{-8pt}
\setlength\tabcolsep{4pt}
\centering
\footnotesize
\begin{tabular}{C{0.2\columnwidth}C{0.2\columnwidth}C{0.2\columnwidth}C{0.2\columnwidth}}
Standard accuracy & \granola{} accuracy & \% of examples  &  BLEURT score \\ \toprule
\checkmark & \checkmark
   & 49.5 & 0.83 \\
\rowcolor{gray!20}
\ding{55} & \checkmark & 5.6 & 0.28 \\
 \checkmark & \ding{55}
 & 0.0 & - \\
\rowcolor{gray!20}
\ding{55} & \ding{55} & 44.9 & 0.26 \\
\hline
\end{tabular}
\caption{Mean BLEURT score for PaLM 2-L with greedy decoding on \graneq, stratified by standard accuracy and \granola~accuracy. 
}
\label{table:bleurt_scores}
\end{table}
\vspace{-0.5em}

\paragraph{Answer annotation in QA datasets.}
QA benchmarks, e.g. Natural Question \cite{kwiatkowski2019natural}, often have multiple answers per question, which may inadvertently include multi-granularity answers. \citet{min2020ambigqa} consider the problem of ambiguous questions, proposing question re-writing to resolve ambiguity. \citet{si2021s} mine answer aliases from a KG and use them to perform ``answer expansion'' to increase the lexical matching score. Our approach is similar but goes one step further, using the KG and LLMs to add multi-granularity answers vs. simply using aliases.

\paragraph{Granularity-driven evaluation.} 
Granularity of model responses has been evaluated in the context of open-domain chatbots, where informativeness plays a crucial role in building engaging dialogue agents. \citet{adiwardana2020towards, thoppilan2022lamda} evaluate granularity, but their focus is on conversational language rather than knowledge evaluation.
\citet{huang2022can} use WikiData to form masked token prediction tasks, such as \nl{Toronto is located in [MASK]}, and test whether pretrained models have a preference for more specific completions (e.g. \nl{Ontario} vs \nl{Canada}). 
Their approach only accommodates single-token predictions and their evaluation covers smaller models. Conceptually, while their objective is to encourage specific answers, we use granularity to perform more faithful evaluation of LM's knowledge and factuality.

\paragraph{Punting.} Abstaining from answering questions 
is a popular approach for improving factuality \cite{kadavath2022language, kuhn2023semantic, yoshikawa2023selective, chen2023adaptation, zhang2023r}. Our approach is motivated by the observation that punting may be overly aggressive; when the model has low confidence in a specific answer but is confident in a coarser answer, outputting the coarser answer is preferred over refusing to answer.

\section{Conclusion}



We highlight a prominent source of factuality errors in modern LMs: generating more detailed responses than their knowledge can support. Using a new QA benchmark, \graneq{}, with multi-granularity answers, and a novel decoding algorithm, \drag{}, we show that taking the answer granularity level into account 
leads to a dramatic increase in model accuracy. In Appendix \ref{appendix:discussion} we discuss various directions for future work.

\section*{Limitations}

Technically, our approach for enriching an existing QA benchmark with multi-granularity answers relies on  extracting entities from the original QA pair and matching them to their KG entry. In less-structured datasets this step may be more involved -- for example, if the surface form of the entity name differs between the dataset and the KG. 

On a more conceptual level, a faithful evaluation of the knowledge of LLMs may also require distinguishing between correct answers based on true knowledge, as opposed to mere educated guesses. This is an issue with QA evaluation in general -- but is especially relevant in our setting, since coarser answers are easier to guess correctly. For example, in the question \nl{Where was [X] born?}, one could guess \nl{Russia} if X is a Russian-sounding name (whereas correctly guessing the city X was born in is less likely). This may require additional information (in the form of providing additional information such as reasoning or evidence) but also relates to how one defines knowledge.

Other than that, our work was demonstrated on a set of large-but-specific LMs from the PaLM model family. Further expanding the study to a wider range of models may also be compelling, but beyond the scope of this work.



\bibliography{anthology,custom}
\bibliographystyle{acl_natbib}

\newpage
\appendix

\section{Directions for future work}
\label{appendix:discussion}

\paragraph{Question perturbations.}
Our approach for generating multi-granularity answers relied on abstractions. A complementary approach would 
modify the question rather than its answer, e.g., altering the question \nl{When was Mark Bils born?}  to \nl{In what year was Mark Bils born?}. 
Such question perturbations could also be coupled with our entity abstraction perspective to generate more broad questions 
like \nl{When was a professor from University of Rochester born?}. Another direction considers generating more specific questions to address knowledge gaps \citep{rabin-etal-2023-covering}. However, question perturbations may create new answers and thus would require more complex evaluation.

\paragraph{Improving \drag{}.}
The two stages of \drag{} -- sampling candidate responses, and response aggregation -- could be improved to yield better granularity adjustment.
For example, it is possible to replace regular temperature sampling \cite{ackley1985learning} with other sampling strategies that may perform better \cite{wang-etal-2022-self, freitag2023epsilon, bertsch2023s}. 
Additionally, better aggregators could improve downstream task performance.

\paragraph{Response granularity fine-tuning.}
While this work focused on improving factuality at inference time, it is interesting to explore fine-tuning with response granularity in mind. For example, \drag{} can be used as a reward model for supervised or RLHF finetuning to encourage models to learn how to tailor the their response granularity to their parametric knowledge or the preceding context.

\section{Additional figures}
\label{appendix:figures}

\begin{figure}[H]
    \centering
    \includegraphics[width=0.5\textwidth]{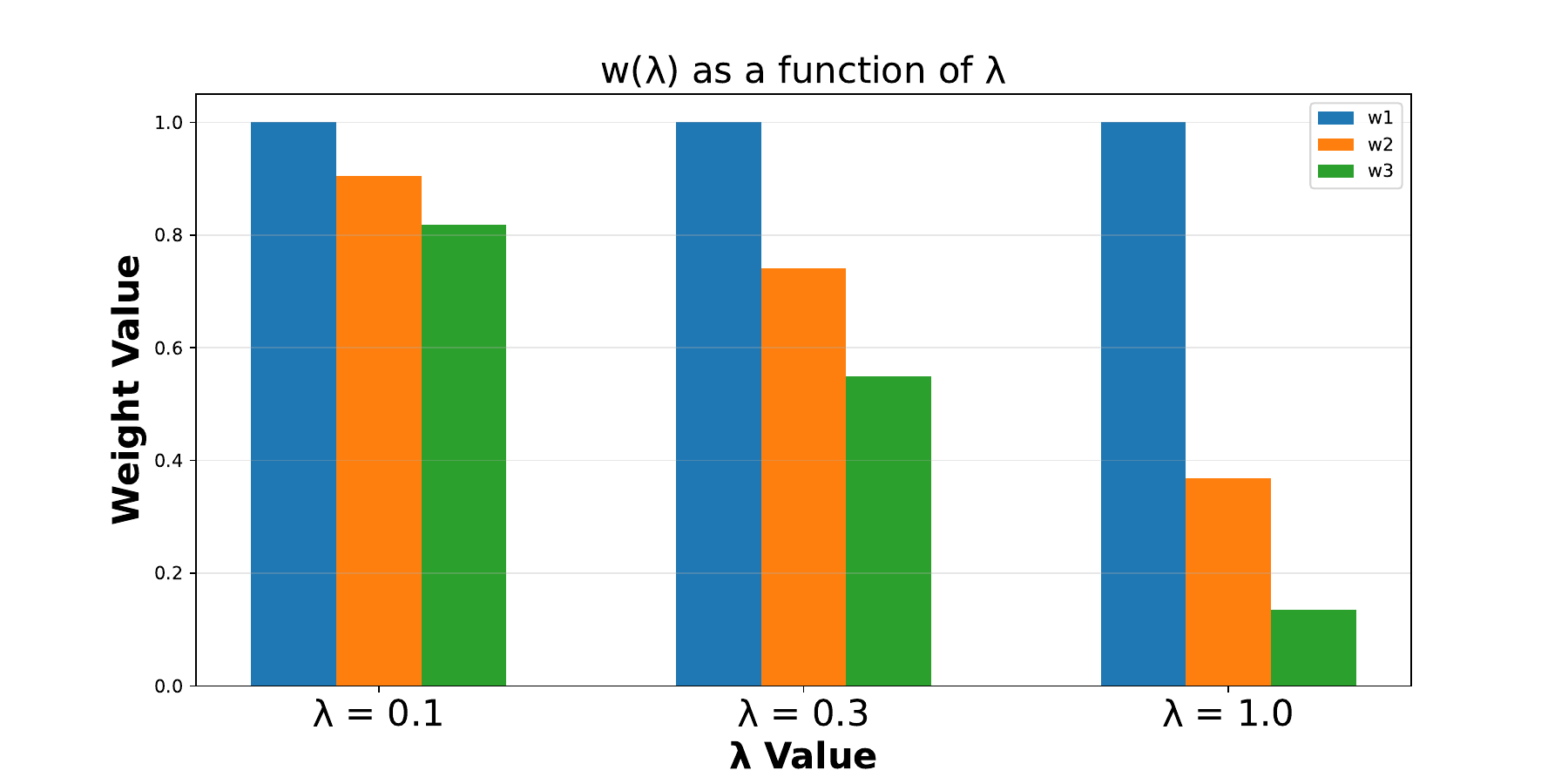}
    \caption{Letting $w_i = \exp(-\lambda (i-1))$, we show how $w_1, w_2, w_3$ behave for different values of $\lambda$.}
    \label{fig:weights}
\end{figure}

In Figure \ref{fig:weights} we show how the parameter $\lambda$ impacts the scores used to evaluate informativeness (see Definition \ref{def:granola_metrics}).

\section{Full details: Constructing \graneq}
\label{appendix:granola-eq}

In this section we detail our process of enriching \eq{}  with multi-granular answers.

\paragraph{Entity Questions.}
\abbreviatedeq{} \cite{sciavolino2021simple}  is a entity-rich QA dataset. It
was created by selecting 
24 common relations from Wikidata and converting fact (subject, relation, object)
triples into natural language questions using manually defined templates.  We restrict our attention to the test split of \eq. In the original \eq{} some of the relations have answers that are already coarse (e.g. \nl{in which continent is country [X]}). We  thus filter examples belonging to such relations.  See Table \ref{table:eq_relations} for the full list of \eq{} and \graneq{} relations.
Additionally, for simplicity, we only keep rows with a unique ground truth example in the original \eq{} dataset. 

\newcolumntype{C}[1]{>{\centering\arraybackslash}p{#1}}

\begin{table}[h]
\footnotesize
\centering
\begin{tabular}{|p{3cm}|C{1.2cm}|C{1.2cm}|}
\hline
\textbf{Template}                            & \textbf{Relation} & \textbf{Included?} \\ \hline
Which country is [X] located in?             & P17               & \ding{55}          \\ \hline
Where was [X] born?                          & P19               & \checkmark         \\ \hline
Where did [X] die?                           & P20               & \checkmark         \\ \hline
Who is [X] married to?                       & P26               & \checkmark         \\ \hline
Which continent is [X] located?              & P30               & \ding{55}          \\ \hline
What is the capital of [X]?                  & P36               & \ding{55}          \\ \hline
Who is [X]'s child?                          & P40               & \checkmark         \\ \hline
Who is the author of [X]?                    & P50               & \checkmark         \\ \hline
Where was [X] educated?                      & P69               & \checkmark         \\ \hline
What kind of work does [X] do?               & P106              & \ding{55}          \\ \hline
Who founded [X]?                             & P112              & \checkmark         \\ \hline
Who owns [X]?                                & P127              & \checkmark         \\ \hline
Where is [X] located?                        & P131              & \checkmark         \\ \hline
What type of music does [X] play?            & P136              & \ding{55}          \\ \hline
Where is the headquarter of [X]?             & P159              & \checkmark         \\ \hline
Who was [X] created by?                      & P170              & \checkmark         \\ \hline
Who performed [X]?                           & P175              & \checkmark         \\ \hline
Which company is [X] produced by?            & P176              & \checkmark         \\ \hline
What music label is [X] represented by?      & P264              & \checkmark         \\ \hline
Where is [X] located?                        & P276              & \checkmark         \\ \hline
Which language was [X] written in?           & P407              & \ding{55}          \\ \hline
What position does [X] play?                 & P413              & \ding{55}          \\ \hline
Which country was [X] created in?            & P495              & \ding{55}          \\ \hline
\end{tabular}
\caption{List of \eq{} and \graneq{} relations.}
\label{table:eq_relations}
\end{table}

\paragraph{Obtaining WikiData descriptions for entities.} The entity extraction stage is simple since the entity location is encoded in the template. For each QA pair we extract two entities: the question entity and the subject entity, and then look these up in WikiData to obtain a WikiData qid for each entity. We then  use the qid to obtain a free text description of the original entity. 

\begin{table}[h]
\footnotesize
\centering
\begin{tabular}{|l|p{5cm}|}
\toprule
\textbf{qid}  & \textbf{description} \\ \hline
\toprule
Q64 & federated state, capital and largest city of Germany \\ \hline
Q142659 & census-designated place in Holmes County, Ohio \\ \hline
Q524646 &  town in Massachusetts \\ \hline
Q614184 &  town in Maryland, United States \\ \hline
\end{tabular}
\caption{QID matches and descriptions for the free text entity \nl{Berlin}.}
\label{table:qid_disam}
\end{table}

\paragraph{QID disambiguation.}
In approximately 30\% of the cases, there are multiple potential qid matches for the same entity (see Figure \ref{table:qid_disam} for an example). We use a simple heuristic for performing 
disambiguation: we select the qid with the smallest value.  

\begin{table}[h]
\footnotesize
\centering
\begin{tabular}{|p{2cm}|p{1.5cm}|p{2cm}|}
\hline
\textbf{Question} & \textbf{QID} & \textbf{QID Description} \\ \hline
Who is the author of Enduring Love?  & Q129813 & 2004 film by Roger Michell \\ \hline
Who performed Orbit? & Q2367904 & historical motorcycle manufacturer \\ \hline
Who is the author of Hollywood?  & Q34006 & neighborhood in Los Angeles, California, United States \\ \hline
\end{tabular}
\caption{When QID disambiguation chooses the incorrect entity, the failure is typically evident since the extracted description (rightmost column) does not semantically match the question.}
\label{table:qid_disam_failures}
\end{table}

\paragraph{Data cleaning.} 
There are two sources of noise in the above automatic process: incorrect extracted descriptions (this may occur when there are multiple WikiData entries for the same entity name, and our disambiguation procedure selects the wrong one) and errors in the LLM generated answers. Thus, to ensure the data is of high quality, we apply several automatic cleaning operations. First, we remove rows containing descriptions that are likely to be erroneous. We utilize the observation that when the QID disambiguation heuristic fails and the wrong QID is selected, this failure will typically be evident 
from the fact that the extracted description is not semantically consistent with the question; see Table \ref{table:qid_disam_failures} for concrete examples. To remove these examples we score each example for how \emph{consistent} the extracted descriptions are with the original questions, and remove examples for which this predicted score exceeds $0.5$. Specifically, we prompt an LLM (with 5 few-shot demonstrations of the intended behaviour) to determine whether the description is consistent (`Yes' or `No'), and we determine the score as the fraction of `No' responses (sampled at unit temperature).  This process ends up removing 1409 examples (or 9.5\% of the dataset). As a second data cleaning step, we remove rows with missing \granola~answers or duplicated answers. Finally, we remove \granola~answers from a list of hard-coded responses that we define as trivial (such as ``person'', ``university'', etc). In total, these steps affected 2378 rows (or 16\% of the dataset). 

\section{Prompts}
\label{appendix:prompts}

Table \ref{tab:algo_prompts} details the prompts used for baseline algorithms. Table \ref{tab:proc_prompts} details the prompts used for the response aggregation sub-routine in \drag{} and for generating multi-granular answers to create \graneq.

\begin{table}[h]
\footnotesize
\centering
\begin{tabular}{|p{2.5cm}|p{4.5cm}|}
\Xhline{1pt}
\textbf{Baseline} & \textbf{Prompt} \\
\Xhline{1pt}
\vanilla & 
\begin{minipage}[t]{4cm}
\vspace{0.2em}
Question: \{question\} \\ Answer: 
\vspace{1em}
\end{minipage}
 \\
\hline
\idk & 
\begin{minipage}[t]{4cm}
\vspace{0.2em}
You will be given a question. Answer the question, or output IDK.
Question: \{question\} \\ Answer: 
\vspace{1em}
\end{minipage}\\
\hline
\uidk & 
\begin{minipage}[t]{4cm}
\vspace{0.2em}
You will be given a question. Answer the question, or, if you are not certain of the answer, output IDK. \\
Question: \{question\} \\ Answer: 
\vspace{1em}
\end{minipage}
\\
\hline
\agg & 
\begin{minipage}[t]{4cm}
\vspace{0.2em}
You will be given a question. Answer the question at a level of granularity that fits your uncertainty, or output IDK. \\
Question: \{question\} \\ Answer: 
\vspace{1em}
\end{minipage}\\
\Xhline{1pt}
\end{tabular}
\caption{Prompts used in the baselines evaluated in \S\ref{section:experiments}.}
\label{tab:algo_prompts}
\end{table}

\begin{table}[h]
\footnotesize
\centering
\begin{tabular}{|p{1.5cm}|p{5.5cm}|}
\Xhline{1pt}
\textbf{Procedure} & \textbf{Prompt} \\
\Xhline{1pt}
Forming multi-granular answers & 
\vspace{0.2em}
\begin{minipage}[t]{5.4cm}
You will be given a pair of question and answer.
You will also receive some additional description about the entity in the question and the entity in the answer.
\\Your task is to write NEW ANSWERS for the original question at various levels of granularity.
Number these answers starting from 1 (with 1 being the most fine grained answer -- the original answer), and
larger indices corresponding to coarser answers.
\\The idea is that someone might not know the answer at the most fine-grained level, but perhaps know the answer at coarser levels.
\\Important: STOP generating answers BEFORE you reach trivial answers.
For example, given the question "who wrote the book X", answers such as "a writer" or "a person" are considered trivial, as these are completely uninformative and can be guessed even without knowing what X is.
\\In your answers, use the format '1:: answer', etc.
\end{minipage}
\vspace{1em}
\\
\hline
Response Aggregation & 
\begin{minipage}[t]{5.4cm}
\vspace{0.2em}
You will be given a list of responses; replace them with the most specific answer that is still consistent with all the original responses.
If the responses have nothing meaningful in common with respect to the question, output IDK.\\Here are some examples:\\ \\ Question: Where was [X] born?\\
Responses:
\\- Hamburg
\\- Hamburg
\\- Bonn
\\- Berlin
\\Correct aggregated answer: Germany
\\Incorrect aggregated answer: Hamburg
\\Explanation: These are all different cities in Germany. Hamburg is not a correct aggregation, since it is not consistent with other responses, such as Berlin or Bonn.
\\
\\Question: When was [X] born?
\\Responses:
\\- February 1, 1937
\\- November 20, 1937
\\- January 1937
\\Correct aggregated answer: 1937
\\Incorrect aggregated answer: November 1937
\\Explanation: These are all dates in 1937. 
\vspace{1em}
\end{minipage}
\\
\hline
\end{tabular}
\caption{Prompts used in \granola~related procedures in the paper.}
\label{tab:proc_prompts}
\end{table}

\section{Additional Results for \S\ref{section:experiments}}
\label{appendix:results}
\newcolumntype{R}{>{\raggedleft\arraybackslash}X}

In this section we include supplementary results from \S\ref{section:experiments}.

\begin{figure}
    \centering
    \includegraphics[width=0.5\textwidth]{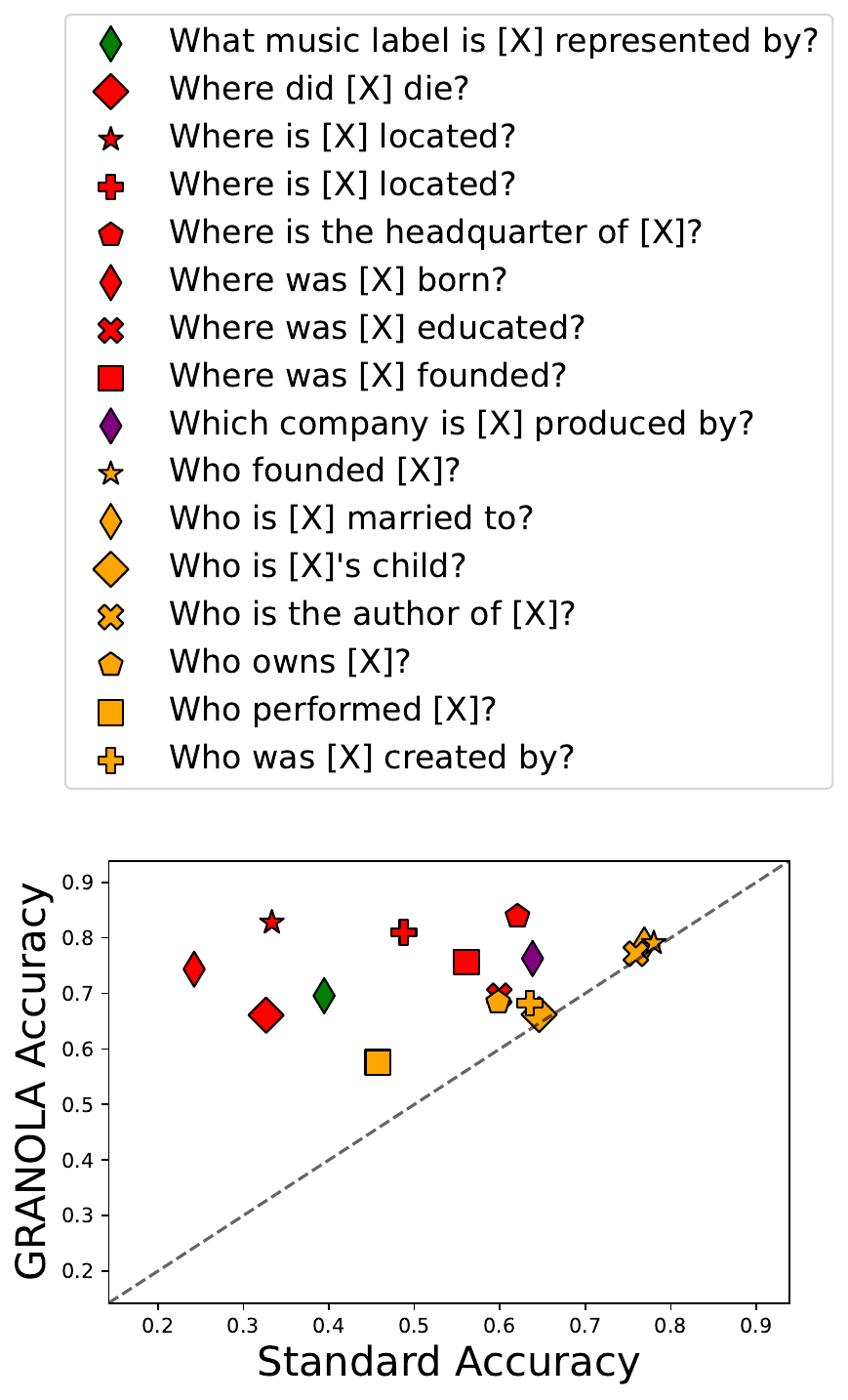}
    \caption{Standard Accuracy (x-axis) vs GRANOLA accuracy (y-axis), stratified by relation, for \drag{} (PaLM 2-L).}
    \label{fig:acc_vs_granola_acc_by_relation}
\end{figure}

\begin{figure*}[t]
    \centering
    \includegraphics[width=1.0\textwidth]{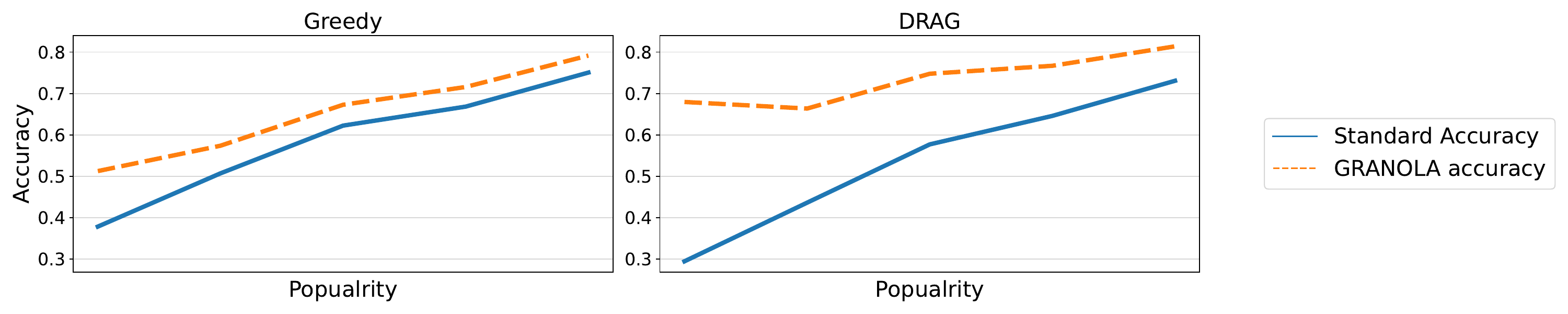}
    \caption{Accuracy (y-axis) vs Entity Popularity (x-axis) for two algorithms: Greedy (left) and \drag~(right). The underlying model is PaLM 2-L. We see that the knowledge evaluation gap is evident for \drag. The behaviour for PaLM 2-M is identical, except the absolute numbers are smaller.}
    \label{fig:acc_vs_pop_340b}
\end{figure*}

\begin{table*}[t]
\footnotesize
\centering
\begin{tabularx}{\textwidth}{|X|X|X|X|X|X|X|}
\hline
\textbf{Question} & \textbf{Original GT Answer} & \textbf{Candidate Answer} & \textbf{Matched GRANOLA Answer} & \textbf{BLEURT Score (GT vs. Candidate)} & \textbf{F1 Score (GT vs. Candidate)} &  \textbf{F1 Score (GRANOLA vs candidate)}\\ \hline
What is Aline Brosh McKenna famous for? & 27 Dresses & screenwriter & being a screenwriter & 0.04 & 0.00 & 0.67 \\ \hline
Where did Tilly Armstrong die? & Carshalton & London & London Borough of Sutton & 0.05 & 0.00 & 0.40 \\ \hline
Where is the headquarter of Guildhall School of Music and Drama? & Barbican Centre & London & City of London & 0.06 & 0.00 & 0.50\\ \hline
Where is Battersea Park located? & Battersea & London & London & 0.06 & 0.00 & 1.00\\ \hline
\end{tabularx}
\caption{Examples from \graneq~where \granola~accuracy disagrees with standard metrics (lexical matching and semantic matching to the original GT answer). The examples were obtained by filtering for example with low F1 score to the original GT answer but high F1 score to the matched \granola~answer, and then sorting by BLEURT scores in ascending order. I.e., they correspond to the points in the top-left corner of Figure \ref{fig:bleurt}.}
\label{tab:examples_mismatch}
\end{table*}

\begin{figure}[h]
    \centering
    \includegraphics[width=0.5\textwidth]{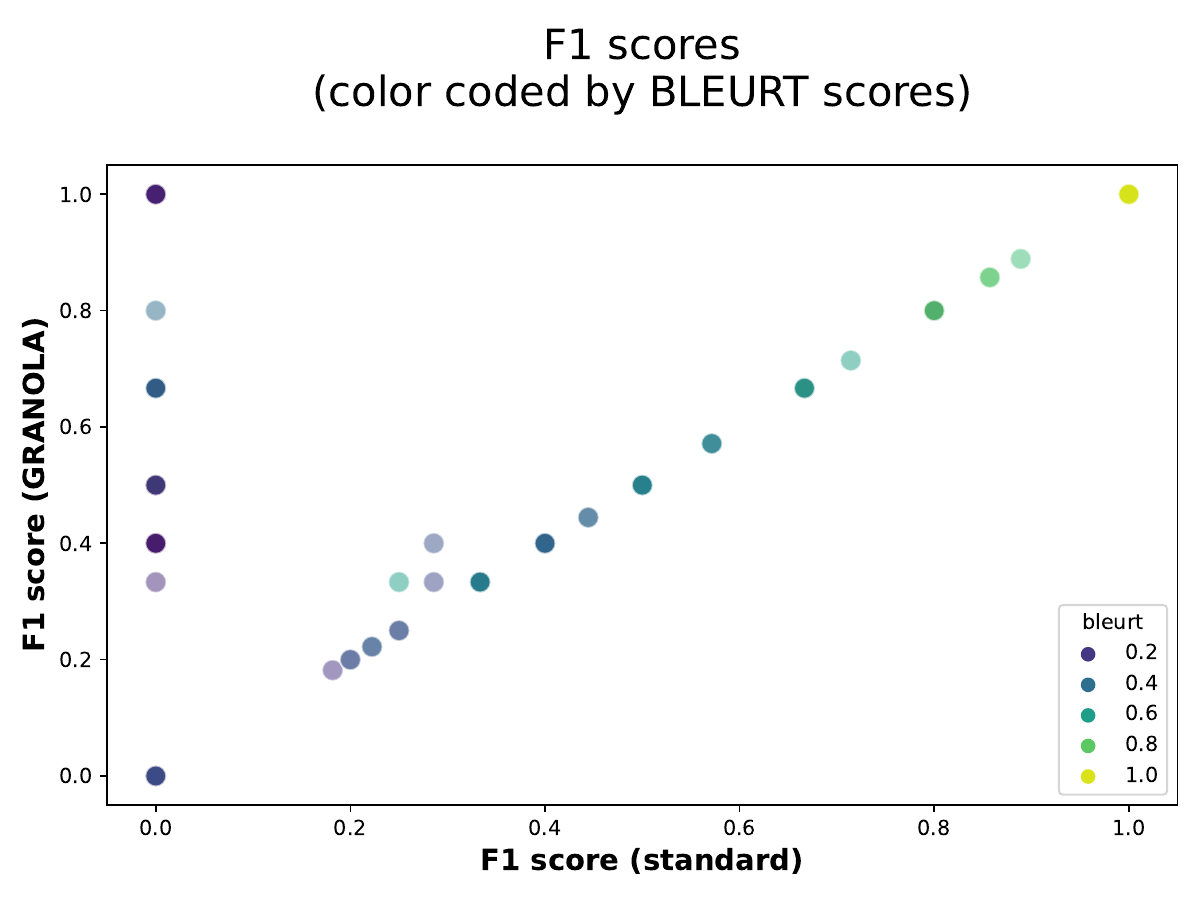}
    \caption{The relationship between standard F1 score (x-axis), \granola~F1 score (y-axis) and BLEURT score \cite{sellam2020bleurt} (color) computed between the original gt answer the candidate answer. The underlying model is Greedy (PaLM 2-L). The figure demonstrates that while there is a strong correlation between standard F1 score and BLEURT scores, this correlation fails specifically for the subset of examples for which \granola~accuracy disagrees with standard accuracy. See Table \ref{table:bleurt_scores} for a quantitative version of this plot. This demonstrates that BLEURT scores can not serve as a replacement for \granola~labels. }
    \label{fig:bleurt}
\end{figure}

\end{document}

%% file: abstract.tex
Factual questions can typically be answered correctly at different levels of granularity. For example, both \nl{August 4, 1961} and \nl{1961} are correct answers to the question \nl{When was Barack Obama born?}. Standard question answering (QA) evaluation protocols, however, do not take this into account explicitly and instead compare a predicted answer against reference answers of a single granularity level. 
In this work, we propose \granola{} QA, a novel evaluation setting where a predicted answer is evaluated in terms of accuracy and informativeness against a set of multi-granularity answers. We present a simple methodology for enriching existing datasets with multi-granularity answers, and create \graneq{}, a multi-granularity version of the \eq{} dataset.\footnote{The data can be found at \url{https://huggingface.co/datasets/google/granola-entity-questions}.}
We evaluate models using a range of decoding methods on \graneq{}, including a new algorithm called Decoding with Response Aggregation (\drag{}), that is geared towards aligning the answer granularity with the model's uncertainty. Our experiments show that large language models with standard decoding methods tend to generate specific answers, which are often incorrect. In contrast, when evaluated on multi-granularity answers, \drag{} yields a nearly 20 point increase in accuracy on average, which further increases for rare entities, revealing that standard evaluation and decoding schemes may underestimate the knowledge encapsulated in language models.